\documentclass[preprint,a4paper]{article}
\usepackage[margin=1.25in]{geometry}
\usepackage[english]{babel}
\usepackage{amsmath}
\usepackage{amsfonts}
\usepackage{graphicx}
\usepackage{subfig}
\usepackage{bm}

\begin{document}

\title{\bf A Bayesian methodology for localising acoustic emission sources in complex structures}	
        
\author{M.R.\ Jones, \ 
        T.J.\  Rogers,\ \ 
        K.\ Worden\, \ 
        E.J.\ Cross \vspace{0.2cm}\\
        Dynamics Research Group, \\
        Department of Mechanical Engineering,\\ University of Sheffield, Mappin Street, S1 3JD, UK
	   }
	\date{}
    \maketitle
	\thispagestyle{empty}

\maketitle

\section*{Abstract}

In the field of structural health monitoring (SHM), the acquisition of acoustic emissions to localise damage sources has emerged as a popular approach. Despite recent advances, the task of locating damage within composite materials and structures that contain non-trivial geometrical features, still poses a significant challenge. Within this paper, a Bayesian source localisation strategy that is robust to these complexities is presented. Under this new framework, a Gaussian process is first used to learn the relationship between source locations and the corresponding difference-in-time-of-arrival values for a number of sensor pairings. As an acoustic emission event with an unknown origin is observed, a mapping is then generated that quantifies the likelihood of the emission location across the surface of the structure. The new probabilistic mapping offers multiple benefits, leading to a localisation strategy that is more informative than deterministic predictions or single-point estimates with an associated confidence bound. The performance of the approach is investigated on a structure with numerous complex geometrical features and demonstrates a favourable 
performance in comparison to other similar localisation methods.

\section{Introduction}

As the demand for more robust and intelligent structural health monitoring (SHM) systems increases, there is interest in not only detecting the presence of damage within a structure, but also identifying its location. By localising the damage, an operator is provided with a greater level of insight into the system, enabling more informed maintenance decisions to be made which will subsequently reduce operation and maintenance costs. 

A prominent feature across many damage localisation strategies is the capture and use of acoustic emission (AE), a phenomenon generated in physical media by mechanical mechanisms such as plastic deformation, friction and crack-tip progression. When an AE event occurs, it results in the release of a small packet of sound energy. Such energy propagates through the material as a high-frequency stress wave, which can then be captured in a digitised manner through the use of a piezoelectric transducer. The basic approach to acoustic emission localisation, which is often referred to as the Time Of Arrival (TOA) method, operates on the premise of observing/detecting an AE source at a number of spatially-distributed sensors. From here, by considering the time of the signal onset at each sensor, a difference-in-time-of-arrival 
(dTOA) value between each sensor pair can be calculated. First proposed by 
Tobias \cite{tobias1976acoustic}, dTOA values were used to produce a hyperbola of potential source locations for each sensor pair. By identifying the intersection of the hyperbolae, an estimate for the origin of the source location was obtained. The TOA method can also be posed as an optimisation problem; in this sense, a source location can be estimated by minimising the difference between each recorded dTOA and a calculated value originating from a trial source 
position. Despite being simple to implement, the TOA approach relies on two assumptions: firstly, that there exists a direct path between the source and sensors, and secondly, that the wave velocity profile is uniform in all directions. However, for anisotropic media such as composite materials, or components 
that contain geometrical features such as thickness changes, bolts and holes, these assumptions become largely invalid. Given that the majority of engineering components will contain at least one of these inhomogeneities, there has been a significant amount of literature dedicated to developing AE localisation strategies that are not constrained to the assumptions detailed above. 

A number of initial studies focussed on developing methods that are robust to anisotropic media, 
incorporating an angular-dependent velocity term into the minimisation 
process \cite{yamada2000lamb,kundu2008locating,nakatani2012impact}. However, knowledge of the wave-velocity profile will not always be available 
\textit{a-priori}. To account for this scenario, approaches that allow a directionally-dependent 
velocity term to be solved in parallel with the event location have also been 
suggested \cite{kundu2012acoustic,ciampa2010new}. In relation to the localisation of AE in structures where a direct propagation path is obstructed, one strategy considered by a number of authors is to define a correlation operator that can quantify the similarity of two AE events through the concept of time-reversal \cite{ing2005solid,park2012impact}. The correlation between a real AE event and a set of artificial events can then be assessed, where the origin of the artificial excitation that returns the highest similarity is determined to be the location of the real source. Ciampa and Meo \cite{ciampa2012impact} 
implemented a similar strategy for localisation in complex geometries, however, by ensuring a diffused wavefield, were able to localise an acoustic impact with only a single sensor. A single-sensor approach was also considered by Ebrahimkhanlou and 
Salamone \cite{ebrahimkhanlou2018single,ebrahimkhanlou2019deep}, training deep neural networks to map a recorded AE waveform to a source location. Although the results shown were promising, the outputs of the neural network were limited to deterministic source location estimates, and therefore provide no metric that quantifies how confident one should be in the predictions. 

For a number of the approaches that have been applied to complex structures, 
there exists a common step across their methodologies; generating and capturing a 
set of artificial AE events across the structure of interest. The use of such a 
technique in the context of AE source location was first suggested as part of the 
Delta $T$ mapping approach developed by Baxter et al.\ \cite{baxter2007delta}. The 
method views AE localisation as a problem of spatial mapping, whereby extracting the onset time of a series of artificial events at a number of distributed sensors, a map that quantifies the expected dTOA information across a structure is constructed for each available sensor-pair. As a real AE event occurs, the recorded dTOA values are then matched to those on the maps, allowing the identification of an estimated source location. The approach has since been applied in a number of challenging environments, such as an aircraft landing gear component \cite{holford2017new}, as well as composite materials \cite{eaton2012acoustic}. An extension to the original methodology was also offered by Al-Jumaili et al.\ \cite{al2016acoustic}, reducing the required level of 
input from an operator. This reduction was achieved by the use of a hierarchical clustering algorithm to automatically discard erroneous training events, as well as the implementation of a minimisation scheme so that a source location estimate could be obtained in an autonomous manner. 

To offer an increased spatial resolution, both the initial and extended Delta $T$ methodologies linearly interpolate between the training grid points to generate the dTOA maps. However, for complex structures, the presence of phenomena such as wave mode conversion and internal reflections will result in a highly nonlinear wavefield. As a consequence of this, a linear interpolation scheme will not be sufficient to capture the complexity of the waveform 
behaviour between grid points. Hensman et al.\ \cite{hensman2010locating} recognised this, and instead proposed the use of Gaussian process (GP) regression, a Bayesian non-parametric technique for nonlinear regression. There, rather than using a minimisation scheme, source localisation is achieved by directly learning an inverse mapping from difference-in-time-of-flight to source location. In addition to providing an improved interpolation strategy, 
a probabilistic output is also returned, where each prediction is associated with a degree of uncertainty - an attractive property of Bayesian solutions. A number of other authors have since adopted various Bayesian approaches as part of AE localisation strategies, including the use of a Markov chain Monte Carlo inference scheme for probabilistically locating sources in a concrete column \cite{schumacher2012toward}, as well as extended and unfiltered Kalman filters for localisation in both isotropic and anisotropic plates 
\cite{niri2012probabilistic,niri2014nonlinear} under noisy TOA measurements. Consideration has also been given by Sen et al.\ \cite{sen2017bayesian} to the particle filter - another example of Bayesian filtering. There, it was demonstrated that AE sources could be located in an isotropic plate whilst incorporating uncertainties such as those arising from material properties and measurement noise. Additionally, Madarshahian et al.\ \cite{madarshahian2019acoustic} proposed the use of probabilistic model selection for identifying AE signal onset times in a Bayesian manner. Whilst these works have returned promising results, localising AE sources from both a probabilistic and general perspective still remains a challenging area of research, particularly for complex structures.

The method presented here, which we term Likelihood of Emission Location (LoEL), will continue in the theme of Bayesian source localisation, and more specifically, extend the use of Gaussian process regression for identifying the location of acoustic emission sources in complex structures. However, unlike in \cite{hensman2010locating}, source localisation is approached through a forward-model solution, 
where source coordinate locations are mapped to difference-in-time-of-arrival values. The likelihood of the emission location is then assessed across the surface of the structure, returning a mapping that quantifies the likelihood an observed dTOA measurement originated from a given location. The new method provides fully probabilistic source location estimates and naturally handles noisy TOA measurements for complex structures. 

\subsection{Outline of proposed methodology} 

This paper presents a probabilistic approach for localising AE sources within complex structures, where under a Bayesian framework, a mapping across a target structure is produced that quantifies the likelihood of the origin of an AE event. In the proposed methodology, the first objective is to employ a Gaussian process to learn the relationship between locations on a test structure and the corresponding dTOA values across a number of sensor pairs. For each unique sensor pair, the 
likelihood that a given dTOA originated from locations across the structure is assessed, generating a contour of potential source locations. A marginal likelihood across all sensors pairs can then be calculated, returning a map of the structure that represents the total likelihood of source location given the model for each sensor pair. Regions of high likelihood are interpreted as being more likely to be the origin of the event, where a 
single-point prediction of the source location can be identified at the location with the highest marginal likelihood. 

In addition to the standard benefits of taking a Bayesian approach, such as predictions that represent a distribution and the ability to naturally incorporate uncertainty, the method offers a number of distinct advantages. By assessing the source location likelihood across the structure, a metric is provided that quantifies how probable is the AE event at a given location. By assessing this likelihood over the surface of the structure, multiple possible damage locations are able to be flagged, as well as providing an 
operator with a greater level of insight than single-point predictions. Secondly, by using coordinate values as inputs to the GP, the noise-free input assumption that standard Gaussian process regression requires is more closely upheld than in the inverse case. There is also the flexibility to deal with targets that have an 
associated level of uncertainty, which given that the difference-in-onset times are often associated with some unknown level of noise (a consequence of factors such as sensor noise and the time of arrival estimation procedure), is a desirable functionality. Lastly, it will be shown that the proposed strategy 
returns a favourable accuracy as the density of the training grid is lowered, particularly in comparison to other leading AE localisation methods. Reducing the number of training measurements will be particularly advantageous for large structures, where generating a fine grid of artificial sources is 
very time-consuming. 

The remainder of the paper proceeds as follows; Section 2 provides an overview of Gaussian process regression theory. Section 3 then introduces the proposed methodology, whilst Section 4 presents a number of results that demonstrate the effectiveness of the method, as well as a discussion. Finally, Section 5 provides an overall conclusion.

\section{Gaussian processes}

Presenting a powerful Bayesian machine learning approach for solving regression problems, Gaussian processes are characterised by their ability to quantify uncertainty on predictions and capture nonlinear relationships in the presence of noisy data. In an engineering world full of noise and uncertainty, this makes GPs particularly relevant for application within SHM, of which other examples can be found in \cite{rogers2020probabilistic,fuentes2014aircraft,civera2017detection}. 

When adopting a Gaussian process framework, one is concerned with mapping a series of $D$-dimensional observations $X = (\mathbf{x}_{\mathbf{i}})_{i=1}^N$ to a set of corresponding vector of targets $\mathbf{y} =(y_i)_{i=1}^N$ which are assumed corrupted by a zero mean Gaussian white noise $\epsilon$, with variance $\sigma_{n}^2$. Mathematically, this equates to modelling functions of the form,

\begin{equation}\label{eq:3.1}
y_i = f(\mathbf{x_i}) + \epsilon, \hspace{30pt} \epsilon \sim \mathcal{N}(0,\sigma^2_n).
\end{equation}

\noindent It is perhaps most intuitive to view the Gaussian process as a distribution over functions, where each realisation is a potential function generated by that GP. In this sense, the GP acts as a prior over $f(\mathbf{x_i})$, and can be defined by,

\begin{equation}\label{eq:3.3}
f \sim \mathcal{GP} (m(\mathbf{x_i}), k(\mathbf{x_i},\mathbf{x_i}')).
\end{equation}

\noindent The GP is fully specified by its mean function $m(\mathbf{x_i})$, and covariance (kernel) function $k(\mathbf{x_i},\mathbf{x_i}')$. Following the standard convention of the machine learning community \cite{williams2006gaussian}, the mean function will be set equal to zero and is therefore omitted from this point onwards. With regard to the kernel, whilst any positive-definite function is a valid covariance function, there are a number of popular choices such as the squared exponential (SE) and the Mat\'ern class of covariance functions. Although the SE is probably the most widely-used, considering the work of Stein \cite{stein1999interpolation}, who argues that the smoothing assumptions imposed by this kernel are often unrealistic when modelling physical process, the choice of kernel here is from the Mat\'ern family. Specifically, the Mat\'ern 3/2 kernel is chosen following the demonstration of its ability to perform well with engineering data \cite{rogers2019towards}. For two inputs $\mathbf{x_i}$, $\mathbf{x_i}'$, the Mat\'ern 3/2 kernel takes the form,

\begin{equation}\label{eq:3.7}
k_{3/2}(\mathbf{x}_{\mathbf{i}},\mathbf{x}_{\mathbf{i}}') = \sigma_f^2(1+ \frac{\sqrt{3} (||\mathbf{x_i}-\mathbf{x_i}'||)}{l})\exp(\frac{-\sqrt{3} (||\mathbf{x_i}-\mathbf{x_i}'||)}{l}),
\end{equation}  

\noindent where $l$ represents the length scale and $\sigma_f^2$ is defined as the signal variance. As the GP is being employed in a spatial modelling context that is asymmetrical about each dimension (see section 3 for details), it makes sense to switch to an anisotropic kernel where a unique length scale is defined for each input dimension. To transform equation (\ref{eq:3.7}) from an isotropic representation, the kernel can be represented as,

\begin{equation}\label{eq:3.10}
k_{3/2}(\mathbf{x}_{\mathbf{i}},\mathbf{x}_{\mathbf{i}}') = \sigma_f^2(1+ \sqrt{3}r)\exp(-\sqrt{3}r),
\end{equation}  

\noindent where

\begin{equation}\label{eq:3.11}
r = \sum_{d=1}^D\frac{||{\mathrm{x}_{id}}-{\mathrm{x}_{id}}'||}{l_d},
\end{equation}

\noindent with $d$ denoting a specific dimension. Having fully defined the form of the Gaussian process, it is then desirable to incorporate the knowledge provided by training data so that predictions can be made at a set of new testing points $X_*$. To achieve this, one can first write the joint distribution of the training targets, $\mathbf{y}$, and the unobserved latent functions, $\mathbf{f}_*$ as,

\begin{equation}\label{eq:3.4}
\begin{bmatrix}
\mathbf{y} \\ \mathbf{f}_*
\end{bmatrix} = \mathcal{N}
\begin{pmatrix}
\begin{bmatrix}
0 \\ 0
\end{bmatrix}, 
\begin{bmatrix}
K(X,X) + \sigma^2_n\mathbb{I} && K(X,X_*) \\ K(X_*,X) && K(X_*,X_*)
\end{bmatrix}
\end{pmatrix}.
\end{equation}

\noindent Utilising standard multivariate Gaussian distribution machinery, the conditional distribution over $\mathbf{f}_*$ can then be obtained \cite{williams2006gaussian} as,

\begin{equation}\label{eq:3.5}
p(\mathbf{f}_*|X,\mathbf{y},X_*) = \mathcal{N}(\mathbb{E}[\mathbf{f}_*], \mathbb{V}[\mathbf{f}_*]),
\end{equation}

\noindent where the posterior mean and variance are,

\begin{align}\label{eq:3.6}
\mathbb{E}[\mathbf{f}_*] &= K(X_*,X)[K(X,X) + \sigma^2_n\mathbb{I}]^{-1}\mathbf{y} \\
\mathbb{V}[\mathbf{f}_*]&= K(X_*,X_*) - K(X_*,X)[K(X,X) + \sigma^2_n\mathbb{I}]^{-1}K(X,X_*).\label{eq:3.61}
\end{align}

\noindent Inspecting equations (\ref{eq:3.10}), (\ref{eq:3.6}) and (\ref{eq:3.61}), it can be seen that there remains a number of terms that must be determined; specifically $l$, $\sigma_f^2$ and $\sigma_n^2$, which can be collated into a single vector $\bm{\theta}$. Defined as the hyperparameters of the model, an estimate of the optimal values, $\hat{\bm{\theta}}$, can be learnt by maximising the log marginal likelihood of the model. As per standard numerical optimisation conventions, this maximisation translates to minimising the negative log marginal likelihood,

\begin{equation}\label{eq:3.8}
  \hat{\bm{\theta}} = \underset{\bm{\theta}}{arg\,min}\{-\log(\mathbf{y}| X, \bm{\theta})\}
\end{equation}

\noindent where, 

\begin{equation}\label{eq:3.9}
\log p(\mathbf{y}| X,\bm{\theta}) = -\frac{1}{2}\mathbf{y}^T(K(X,X) +\sigma^2_n\mathbb{I})^{-1}\mathbf{y} - \frac{1}{2}\log |(K(X,X)+\sigma^2_n\mathbb{I})| - \frac{N}{2}\log(2\pi).
\end{equation}  

\noindent Generally speaking, a numerical optimisation scheme of choice can then be employed to evaluate equation (\ref{eq:3.8}). The authors' preference here is an example of population-based optimisation - the Quantum-Behaved Particle Swarm (QPSO) \cite{sun2004global}. This approach has the key advantage of being globally convergent \cite{sun2004particle}, and has been demonstrated to perform well in engineering optimisation challenges, of which an example can be found in \cite{worden2018evolutionary}.

\section{Localising AE sources with Gaussian processes}\label{section3}

Having established the mathematical framework for implementing Gaussian process regression, this section will detail the proposed source localisation strategy. The experimental methodology used to demonstrate the approach will firstly be considered, before discussing how a Gaussian process can be used to learn a transformation from grid locations to dTOA values. Finally, the necessary framework for quantifying the source location likelihood will be given. It should be noted that the experimental set-up is identical to that used in \cite{hensman2010locating}; however, for clarity, the key points will be reiterated. 

\subsection{Experimental setup}

\begin{figure}[h]\label{fig:f11}
  \centering
  \subfloat[Test structure schematic, with labelled transducers.]{\includegraphics[scale=0.65]{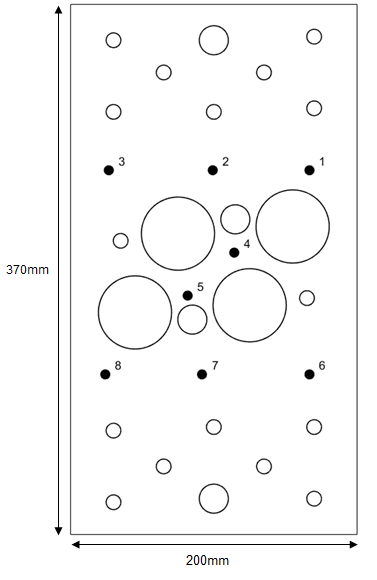}\label{fig:f1}}
  \hspace{0.005em}
  \subfloat[Image of test structure.]{\includegraphics[scale=0.65]{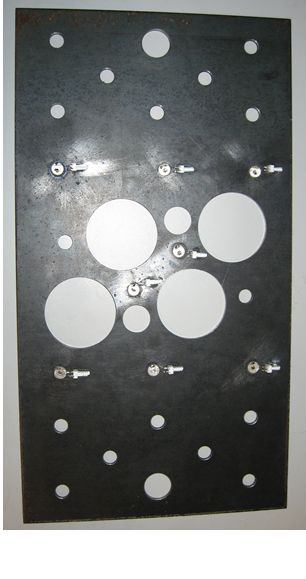}\label{fig:f2}}
  \caption{Complex structure used to demonstrate the methodology. Recreated from \cite{hensman2010locating}.}
\end{figure}

The specimen used to demonstrate the proposed methodology is a 200x370x3mm aluminium sheet, which as shown in Figure 1, contains a series of holes that have been cut through the plate. These non-trivial geometrical features then introduce a number of complex phenomena that result in a challenging environment for performing acoustic source localisation. A grid was first constructed across the surface of the structure, where at each nodal location, a high-power laser pulse was used to artificially generate an acoustic emission source. Eight piezoelectric transducers were mounted to the specimen and used to capture the ultrasonic response to each excitation. The locations of these sensors are indicated by the black circles on Figure \ref{fig:f1}.  

\subsection{Learning process}

Before learning a GP representation, the data set that has been gathered needs to be processed to reflect the relationship that is being modelled. Considering a single artificial AE event, there will be eight captured waveforms, each corresponding to an individual sensor. To extract the difference-in-time-of-arrival information, the arrival time of the waveform at each sensor is first estimated through the standard practice of implementing the Akaike Information Criterion (AIC) method (for more details, see \cite{kurz2005strategies}). It is then trivial to calculate the difference-in-time of arrival across all sensor pairs. Given that the use of eight individual sensors results in 28 unique pairing combinations, for each AE event, there will be a corresponding dTOA vector containing 28 entries. 

\begin{figure}[h]
  \centering
  \subfloat[dTOA values of artifical AE sources, \newline where each circle represents one event.]{\includegraphics[scale=0.525]{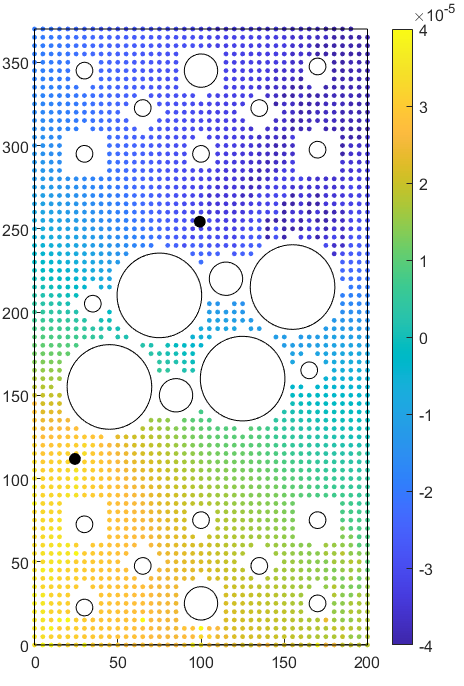}\label{fig:f3}}
  \hspace{0.00005em}
  \subfloat[dTOA learnt by a Gaussian process.]{\includegraphics[scale=0.525]{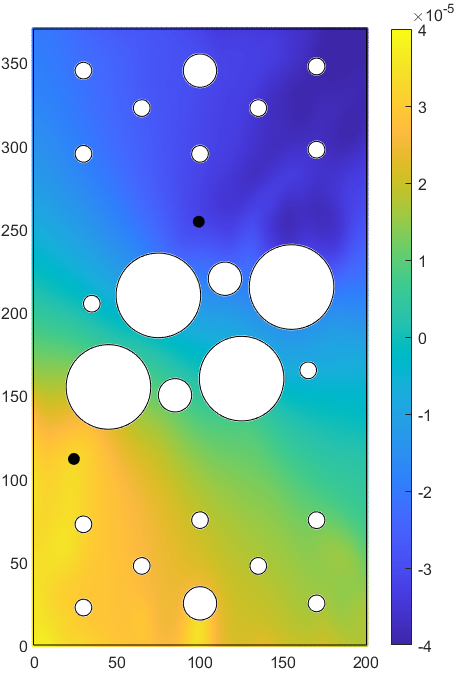}\label{fig:f4}}
  \caption{Spatial maps of dTOA values for sensor pairing 2-8, where the colour bar represents dTOA in seconds.}
\end{figure}

Having established a suitable data set, the process of learning a relationship between source location and dTOA can begin, where a GP representation is learnt for each unique sensor pair. Providing a graphical illustration of this process for sensor pairing 2-8, Figure \ref{fig:f3} represents each of the values used in the training period, whilst Figure \ref{fig:f4} demonstrates a denser mapping that has been learnt by the GP.

\subsection{Source location estimation}\label{section33}

In a standard regression setting, one is interested in capturing a relationship between two sets of variables; those termed the independent or explanatory variables, and those for which the objective is to predict the values of (the dependent variables). Under a Gaussian process framework, the independent variables are generally assigned as the input points, whilst the quantities one wishes to predict are placed on the targets. In the context of source localisation, the desired objective is to estimate the location of an observed AE event, suggesting that a location coordinate be designated as a target with a dTOA value (for example) assigned as an input. Although this direct means of modelling an inverse problem has demonstrated success \cite{hensman2010locating}, it is not reflective of the general approach to regression where the independent variables are assigned as model inputs. This paper avoids such an issue by considering a forward model, where the potential source location coordinates, which are indeed the explanatory ones, are used as model inputs, with the dTOA values assigned as model targets.  

An additional benefit of adopting a forward model is that the underlying assumption in standard GP regression of noise-free inputs \cite{williams2006gaussian} is more closely met than if taking an inverse model approach. Although the AIC method yields a reasonable prediction of the onset times, they are still only estimates of the true values. The dTOA values should, therefore, be interpreted as being corrupted with some level of noise, and therefore if placed on the inputs of the GP, do not uphold the required assumption. This will be particularly pertinent when operating outside of the ideal conditions of the laboratory, where increased levels of external noise will have more of a disruptive influence on the time of flight estimates. In the case of the AE source locations, the user has direct control of the position at which these events are generated. Therefore, providing that the experiment has been set up with good practice in mind (such as proper calibration of positioning devices), it is viable to assume that the noise present on the AE event locations is insignificant relative to that on the dTOA measurements. 

By placing grid coordinates on the input to the GP, the location of a future observed AE event can no longer be estimated solely by the predictive distribution. Instead, having mapped a grid of coordinates onto their corresponding dTOA values via GP regression, it is possible to assess the likelihood of a new set of observed dTOAs for each sensor pair $\mathbf{y}_{obs} = \left\{y_{obs,1},y_{obs,2},\ldots,y_{obs,J}\right\}$ given the learnt functional mapping from the $(x,y)$ coordinates to a measured dTOA in the training data, again for each sensor pair $j=1,2,\ldots,J$. For the set of $J$ models, this means assessing, 

\begin{equation}\label{eq:4.2}
\log p(y_{obs,j}|\mathcal{D},\mathbf{x_*},\mathcal{M}_j) = -\frac{1}{2}\log\mathbb{V}[\mathbf{f_*}_{,j}] - \frac{(y_{obs,j}-\mathbb{E}[\mathbf{f_*}_{,j}])^2}{2\mathbb{V}[\mathbf{f_*}_{,j}]} -\frac{1}{2}\log2\pi,
\end{equation}

\noindent where $\mathcal{D} = \left\{\mathbf{X}_i,y_{i,j}\right\}_{i=1}^N$ is the set of $N$ training pairs where, for each $i$, $\mathbf{X}_i$ is the $(x,y)$ coordinate for that observation and $y_{i,j}$ is the dTOA for observation $i$ at sensor pair $j$. Equation \ref{eq:4.2} gives the likelihood of a newly observed dTOA from an unknown source location given a candidate location $\mathbf{x_*}$ for a given sensor pair, i.e. model $\mathcal{M}_j$. 
Although this likelihood remains conditioned on the hyperparameters, it is assumed that these are now fixed given the training stage of the model and they are not shown to avoid clutter. 

By calculating the exponential of equation (\ref{eq:4.2}) at a set of candidate points across the structure (i.e. for $\mathbf{x_*} = X_*$), the likelihood that the event originated from each point can be computed. Defining this set of candidate locations as the predictive grid, it is clear that the accuracy of using equation (\ref{eq:4.2}) to identify the source origin will be dictated by the density of the grid. When implementing a GP, the main computational burden lies with the inversion of the covariance matrix, scaling as $\mathcal O(N^3)$. Given that this operation is exclusive to the training stage, defining a suitably dense predictive grid will generally not be the dominant factor in computing time. 

\begin{figure}[h]
\centering
\includegraphics[scale=0.41]{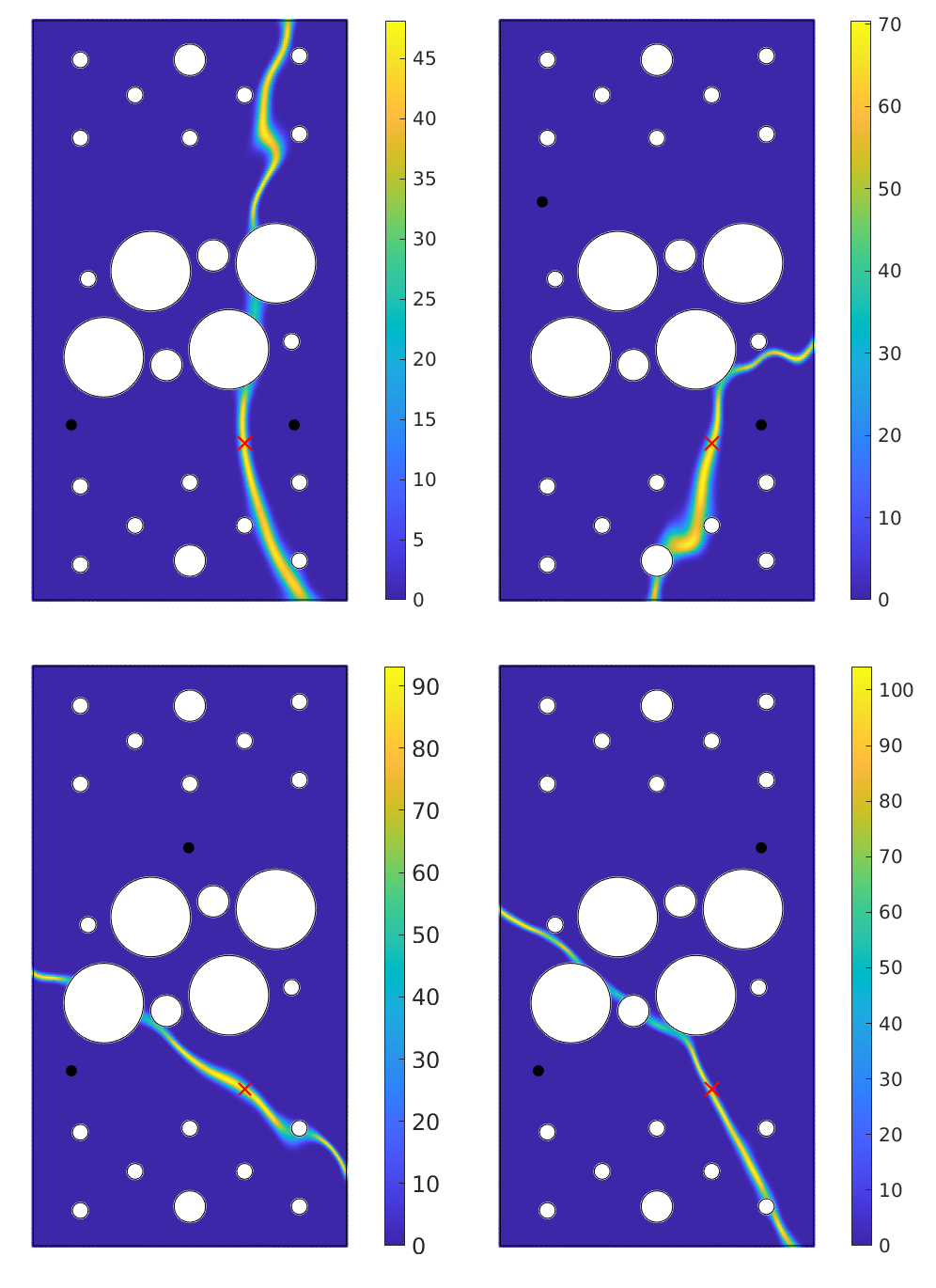}
\caption{Source location likelihood maps corresponding to four individual sensor combinations, where the colour bar represents the likelihood value. The sensor pair used is represented by black circles, whilst the red cross marks the true source location.} \label{fig5}
\end{figure}

To demonstrate the method visually, the data is split into a training and testing set, where each data point consists of the source location and the corresponding dTOA vector for an artificial event. For a single AE event randomly selected from the testing set, Figure \ref{fig5} shows the likelihood that the corresponding dTOA value originated from each location on the predictive grid. The results are shown for four sensor-pairings, where the red cross indicates the true location.

Investigating each of the maps, it can be seen that rather than providing a unique solution, there exist contours of highly probable locations for each sensor pair that will generally intersect at the true location. This is of course expected; acoustic emission theory states that when considering difference-in arrival times of the fastest wave mode to localise a source, providing the source does not sit on a direct path between two sensors, a minimum of three sensors are required to provide a unique solution. To account for this behaviour, it is necessary to marginalise over each of the individual models, which is equal to the following,

\begin{equation}
    p\left(\mathbf{y}_{obs}\;\vert\; \mathcal{D}, \mathbf{x_*}\right) = \int p\left(\mathbf{y}_{obs}\;\vert\; \mathcal{D}, \mathbf{x_*}, \mathbf{\mathcal{M}}\right)p\left(\mathbf{\mathcal{M}}\right) d\mathbf{\mathcal{M}}
    \label{eq:marg_mod}
\end{equation}

\noindent By observing that the likelihood in each sensor pair depends only on the dTOA between those sensors ($y_{obs,j}$) and that the collection of sensor pairs forms a finite and discrete set of possible $\mathbf{\mathcal{M}} = \left\{\mathcal{M}_1,\mathcal{M}_2,\ldots,\mathcal{M}_J\right\}$, the integral in equation \ref{eq:marg_mod} can be transformed into a finite sum.

\begin{equation}
    p\left(\mathbf{y}_{obs}\;\vert\; \mathcal{D}, \mathbf{x_*}\right) = \sum_{j=1}^J p\left(y_{obs,j}\;\vert\; \mathcal{D}, \mathbf{x_*}, \mathcal{M}_j\right)p\left(\mathcal{M}_j\right)
\end{equation}

\noindent Placing an equal importance on all sensor pairs, it is possible to set $p\left(\mathcal{M}_j\right) = 1/J \;\forall\; j \in \left\{1,2,\ldots,J\right\}$. %
Therefore, 

\begin{equation}
    p\left(\mathbf{y}_{obs}\;\vert\; \mathcal{D}, \mathbf{x_*}\right) = \sum_{j=1}^J p\left(y_{obs,j}\;\vert\; \mathcal{D}, \mathbf{x_*}, \mathcal{M}_j\right)
    \label{eq:marg_like}
\end{equation}

\noindent The task of predicting the true source location for the measured emission event can then be framed as maximising the likelihood of the observed dTOAs given the candidate location $\mathbf{x_*}$ and the previously observed training data $\mathcal{D}$. %
In other words to determine, $$ \hat{\mathbf{x}}_* = \underset{\mathbf{x}_*}{arg\,max}\left\{\log p\left(\mathbf{y}_{obs}\;\vert\; \mathcal{D}, \mathbf{x_*}\right)\right\}$$

\noindent where the candidate AE source location $\mathbf{x}_*$ with the highest likelihood of generating the observed dTOAs from the observation $\mathbf{y}_{obs}$ is determined to be the predicted source location, $\hat{\mathbf{x}}_*$. The advantage of marginalising across the possible models (sensor pairs) is that one is not required to choose a best subset of sensors with which to proceed, instead it is possible to capture the information from all possible measurements and combine them in a consistent manner. %
It also allows, if \textit{a-priori} knowledge is available, weighting of different sensor pairs based on their efficacy in identifying source locations through modification of $p\left(\mathbf{\mathcal{M}}\right)$.

\section{Results and discussion}

\begin{figure}
\centering
\includegraphics[scale=0.85]{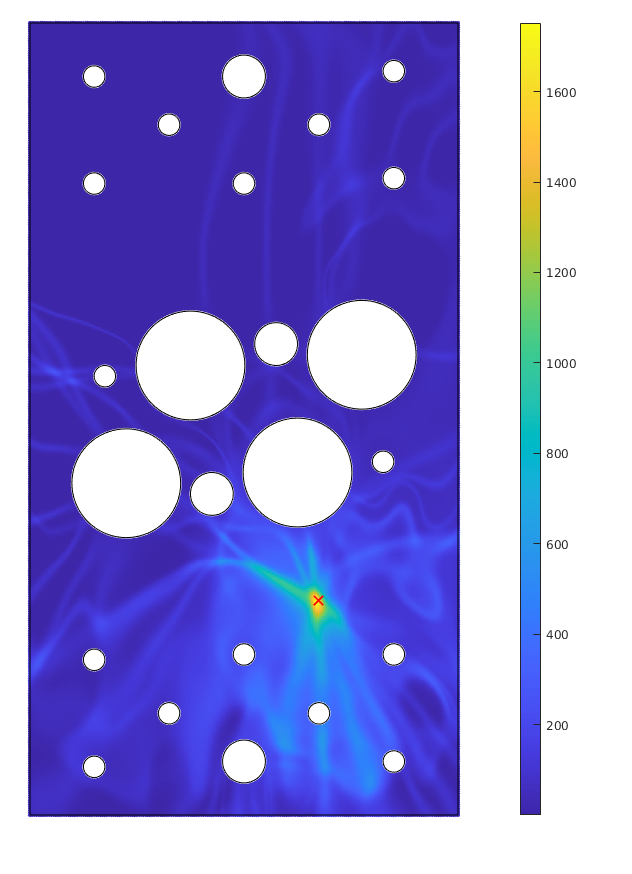}
\caption{Mapping that quantifies the combined source location likelihood of an AE event. The true origin is given by a red cross and the colour bar represents the combined likelihood value.} \label{fig6}
\end{figure}

Following the implementation of the localisation strategy detailed in the previous section, for the same test point used in Section \ref{section33}, Figure \ref{fig6} represents the marginal likelihood across all sensor-pair models that the event originated from locations across the plate. Inspecting Figure \ref{fig6}, it can be seen that there exists a region of high likelihood around the true location, demonstrating that the origin of the AE event has been correctly located. If desired, a single estimate of the origin location can then be made by considering the location on the map that returns the highest marginal likelihood. 

The key advantage of the probabilistic maps is that the source location likelihood can be evaluated at any number of locations across the plate. The user is then flexible to decide on how to proceed. The simplest option is to use the algorithm in isolation, where the maximum likelihood solution can be used to identify the most likely location of the AE source. Alternatively, it is also possible to feed the location likelihoods into a wider risk-based SHM framework, where likelihood values are used to inform the calculation of risk associated with some 
given event \cite{flynn2010bayesian,vega2020variational,hughes150probabilistic}. 

In addition to identifying the highest likelihood event location, the probabilistic source location maps also provide an intuitive visualisation of the associated confidence, which can be used to inform an operator how large a region of the structure may warrant closer inspection. For example, in the case that the algorithm identifies the most likely source location with high certainty, one would expect to see a single compact region of high likelihood, as observed in Figure \ref{fig6}, and therefore requires only a small area of the structure to be inspected. However, where a prediction is made with less certainty, this may be reflected in either a single region of high likelihood becoming more spread, or through the emergence of multiple highly probable locations. To demonstrate a prediction with lower confidence, Figure 5 provides the source location likelihood map for a test point where the region of high likelihood is far more dispersed about the true location than that of the previous example, and indicates that a larger region should be investigated further. In this case of low confidence, the ability to flag an area/number of potential source locations will prove more informative over considering single point estimates with an associated uncertainty bound. 

\begin{figure}
\centering
\includegraphics[scale=0.85]{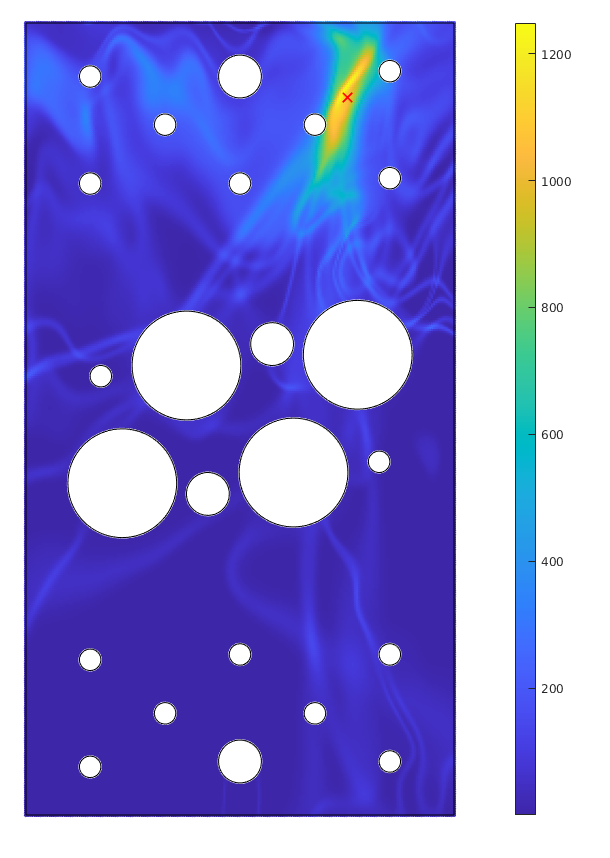}
\caption{Mapping with an increased uncertainty around the location of the AE event. The true origin is given by a red cross and the colour bar represents the combined likelihood value.} \label{fig9}
\end{figure}

\newpage
\subsection{Single location prediction performance}

\begin{figure}[h]
\centering
\includegraphics[width=0.89\textwidth]{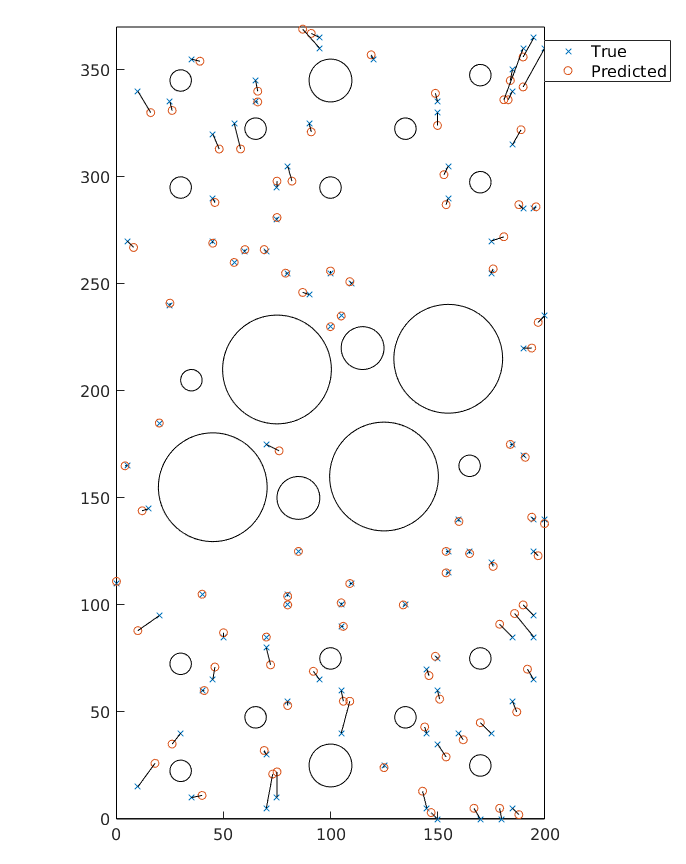}
\caption{Comparison between the predicted and true AE event location for 100 testing events.} \label{fig7}
\end{figure}

Whilst single location predictions are not the sole focus of this work, it is still useful to consider them as a way to assess the general performance of the localisation algorithm. To do this, the predicted location (i.e. the most likely) of each observation in the test set is compared to the true value, with the results illustrated in Figure \ref{fig7}. Upon visual inspection, it can be seen that there is a high level of agreement between the predicted and true location for the majority of points within a test set of 100 events. Assessing the root-mean-squared-error (RMSE) of the test set according to equation (\ref{eq:5.1}) returns a value of 4.57 mm, which given the dimensions of the test structure, demonstrates the accuracy of the proposed approach.

\begin{equation}\label{eq:5.1}
RMSE = \sqrt{\frac{\sum_{x,y}((x,y)_{pred} - (x,y)_{true})^2}{N_{test}}}.
\end{equation}

\subsection{Sensitivity to training grid size}

\begin{figure}[h]
\centering
\includegraphics[width=1\textwidth]{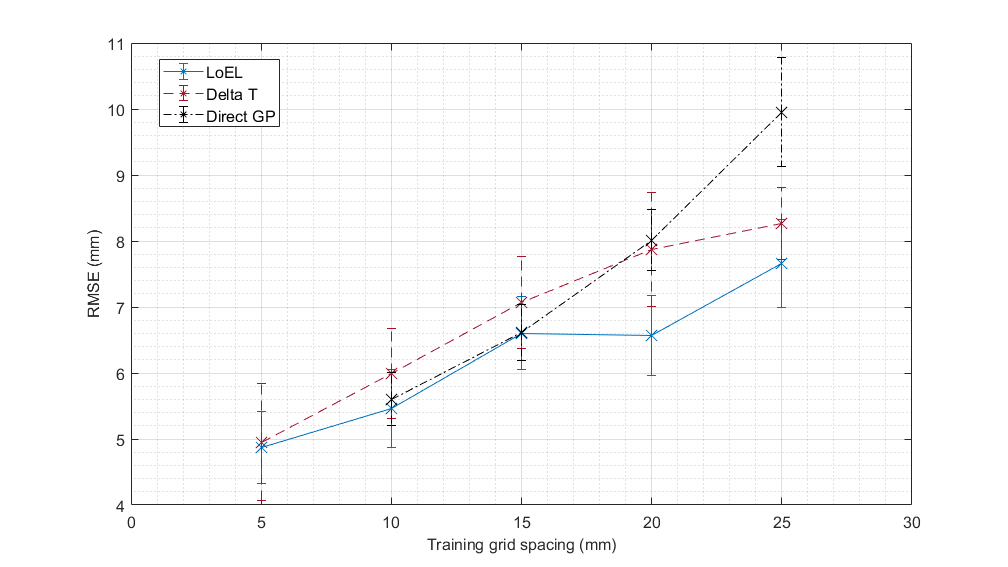}
\caption{Mean and standard deviation of RMSE for 10 test sets as the spacing between the number of training set observations is varied.} \label{fig8}
\end{figure}

One of the main costs associated with implementing a data-driven localisation strategy is the acquisition of training data, requiring artificial AE events to be generated across the test structure. Whilst autonomous data-collection solutions have been suggested \cite{hensman2010locating}, it is still beneficial for the number of training events to be reduced so that less time has to be spent collecting data. To investigate how sensitive the localisation algorithm is to the number of data points used in the training phase, the RMSE of single-point predictions using training sets with different 
grid spacing is investigated, where grid spacing is defined as the distance between each training point in both the $x$ and $y$ direction. For each training grid size, an averaged RMSE across ten testing sets that each consist of 100 randomly selected locations is then assessed, with the results plotted in Figure \ref{fig8}. An obvious first conclusion to make is that as the spatial resolution is increased, which corresponds to reducing the grid spacing, the RMSE decreases. This is entirely expected; a training set representative of a denser grid 
will require less of the feature space to be interpolated over. A more notable observation is that as the grid spacing becomes larger, the corresponding increase in error is of a significantly smaller factor. For example, shifting the grid spacing from 5mm to 20mm - which is a 300\% increase - returns only a 34.4\% rise in error. Given that increasing the grid spacing from 5mm to 20mm would allow the required number of training events to be reduced from 2277 to 154, in many circumstances, the relatively small increase in error could be justified by the considerable amount of time that would be saved during the acquisition stage. Additionally, given that the data points are naively removed in a 
uniform manner, following a more informed training point selection process, such as 
removing less events around geometrical features, would likely see more favourable 
errors as the number of training measurements is reduced. 

\subsection{Comparison to existing methods}

To compare the LoEL method with similar state-of-the-art approaches, the exercise 
completed in the above section is repeated using both the latest variant of the 
Delta $T$ mapping technique (details in 
\cite{al2016acoustic}) and also the method presented by 
Hensman et al.\ \cite{hensman2010locating}, which will be termed the direct GP 
approach. As this is the authors' implementation of both of these methods, the 
training and testing locations will differ to those presented in the results 
of \cite{hensman2010locating}. Additionally, it should be noted that a grid 
spacing of 5mm was not computed for the direct GP due to excessive computational 
demands, which was estimated to be two weeks run time on a standard workstation due to re-training the model for each test set. 

Considering Figure \ref{fig8}, LoEL compares favourably with both Delta $T$ and the direct GP approach, returning a lower RMSE at the majority of the training grid resolutions. An increased predictive performance is particularly evident at the larger grid spacings, where it can be seen that the difference in error between the proposed approach and those for comparison increases. This enhanced interpolative capability can be ascribed to two factors. Firstly, as the behaviour of the acoustic emission waves across the plate is inherently nonlinear, 
the accuracy of the interpolation is dependent on the ability to represent such complexity. In the proposed approach, the choice of kernel function within the GP prior allows nonlinear relationships to be captured, whilst for Delta $T$, the method is limited to fitting a linear regression between known data points. Therefore as training data become more sparse, the assumption 
of a linear functional form will become increasingly restrictive, leading to a worsened interpolative performance. Secondly, as linear interpolation fits exactly through known targets, it is not possible to directly account for the process noise associated with the difference-in-time-of-arrival values. Conversely, under the LoEL approach, the dTOA measurements are considered to be the summation of a 
true underlying value, and some Gaussian noise term, which, for reasons discussed in Section 3.3, is more reflective of the underlying process being modelled. 

For the direct GP, the radial basis function (RBF) kernel (equivalent to an SE) that is implemented is capable of modelling nonlinearities and therefore overcomes the first limitation of Delta $T$. However, the infinitely differentiable property of the RBF kernel likely makes too strong a smoothing assumption on the data. By 
switching to a Mat\'ern kernel, the family of functions believed to represent the 
relationship between spatial location and dTOA is now restricted to those that are $p$ times differentiable (where $p = \nu - 1/2$, with $\nu = 3/2$). This order of differentiability more accurately reflects the underlying physical process being modelled, contributing to an improved spatial interpolation for our method. In relation to accounting for the uncertainty associated with the AE onset times, by mapping from dTOA measurements to a spatial location, the assumption of noise-free inputs required by standard Gaussian process regression is not upheld, whilst the uncertainty associated with the TOA measurements is unable to be incorporated into the model. This will clearly be detrimental to the localisation accuracy, particularly for the two least dense training sets where the error rises sharply in comparison to our method. An additional benefit of the proposed approach over the direct GP is a significant reduction in learning time. For the direct GP, a Gaussian process prior is learnt 
for each grouping of three or more sensors, which for eight individual sensors, corresponds to 438 GPs. In the case of the method proposed here, a GP is trained for each sensor paring, reducing the number of GPs to 28. Although Hensman et al.\ \cite{hensman2010locating} outlines that the same set of hyperparameter values can be used for each array of the same dimensionality, it is not clear how a subset of trial hyperparameter values would be identified in an efficient manner. 

\section{Conclusion}

The work contained within this paper has established a probabilistic framework for 
localising acoustic emission sources in complex structures, offering a number of advantages:

\begin{itemize}
\item The likelihood that an observed AE event originated from a given location is quantified across the surface of the structure, where the location that returns the maximum likelihood identifies the most likely source origin.
\item By using a forward model approach and placing the locations of the training events on the GP inputs, the model is representative of how regression problems are generally treated, as well as more closely upholding the noise-free input assumption than in the case of a direct strategy.  
\item There is the potential ability to flag multiple possible damage locations.
\item The probabilistic mapping offers an intuitive visualisation of the confidence associated with a location prediction, which can assist an operator in deciding how large of an area to inspect.  
\item As the number of data points used in the training phase is reduced, when applied to a complex-like structure, an enhanced interpolative performance is demonstrated in comparison to other similar localisation methods. 
\end{itemize}

\noindent Considering future work, it would be desirable to explore alternative approaches for combining predictions across sensor pairs. For example, a weighting scheme that is able to account for the uncertainty associated with individual predictions will result in a more robust approach, particularly as the number of training measurements are reduced. 

\section*{Acknowledgements}

The authors would like to gratefully acknowledge the support of grant reference numbers EP/S001565/1 and 
EP/R004900/1. Additionally, thanks is offered to James Hensman, Mark Eaton, Robin Mills and Gareth Pierce for their work in acquiring the data set used within the paper. 

\bibliographystyle{IEEEtraN}
\bibliography{References_pm.bib}
\end{document}